\documentclass{article}
\usepackage{spconf,amsmath,graphicx}
\usepackage{amssymb,amsfonts}
\usepackage{multirow}
\usepackage[capitalise]{cleveref}
\usepackage{caption}
\usepackage{subcaption} 
\usepackage{siunitx}
\usepackage{xurl}

\usepackage{flushend}

\title{Estimation of 3D Body Shape and Clothing Measurements from Frontal- and Side-view Images}
\def\correspondingauthor{\sthanks{Corresponding author: sungho.suh@dfki.de}}

\name{Kundan Sai Prabhu Thota $^{1,2}$, \ Sungho Suh$^{1,2}$\correspondingauthor, \ Bo Zhou$^{1,2}$, \ Paul Lukowicz$^{1,2}$}
\address{$^{1}$ Department of Computer Science, TU Kaiserslautern, 67663 Kaiserslautern, Germany\\
	$^{2}$ German Research Center for Artificial Intelligence (DFKI), 67663 Kaiserslautern, Germany}

\begin{document}
%
\maketitle

\begin{abstract}
  The estimation of 3D human body shape and clothing measurements is crucial for virtual try-on and size recommendation problems in the fashion industry but has always been a challenging problem due to several conditions, such as lack of publicly available realistic datasets, ambiguity in multiple camera resolutions, and the undefinable human shape space. Existing works proposed various solutions to these problems but could not succeed in the industry adaptation because of complexity and restrictions. To solve the complexity and challenges, in this paper, we propose a simple yet effective architecture to estimate both shape and measures from frontal- and side-view images. We utilize silhouette segmentation from the two multi-view images and implement an auto-encoder network to learn low-dimensional features from segmented silhouettes. Then, we adopt a kernel-based regularized regression module to estimate the body shape and measurements. The experimental results show that the proposed method provides competitive results on the synthetic dataset, NOMO-3d-400-scans Dataset, and RGB Images of humans captured in different cameras.
\end{abstract}
\begin{keywords}
3D Shape Estimation, Clothing Measurements, Auto-encoder, Silhouette Segmentation, SMPL
\end{keywords}
\section{Introduction}
 Human body shape and clothing measurement estimations are essential to design garments that are a perfect fit for the user. Prior to the surge of online retailing, garment suppliers did not have to worry about the perfect fit while making the garments since the in-store shoppers could physically try on the clothes. 
    While online retailing has brought unparalleled convenience for the customers, the decoupling between the physical presence and the shopping process has challenged the fashion industry to provide good-fitting garments for every customer.
    According to recent surveys \cite{news, barclay}, inconsistent sizing causes a major part of the e-commerce clothing returns, which has led to additional costs for the retailers, landfill waste, and carbon emission.
    A customized fit is particularly difficult due to factors such as regional, brand, and store variances, as well as the customer's body shape changing over time.
    At present most retailers refer to the past successful purchase history. 
    While customers could provide the bust/waist/hip measurements to the online retailers, relying on the custome   rs alone to perform those measurements not only brings inconvenience but also still suffers from the aforementioned problems such as the standard detachment between the customers and the producers.
    Therefore, online retailers must have a solution to obtain reliable body measurements more easily for the customers to improve satisfaction. 

    Earlier works \cite{dibra2017human, ma2018learning, ji2018shape, gai20193d, yan2021silhouette} have proposed 3D body shape and clothing measurement estimation methods on silhouettes of humans or synthetic data. Dibra et al. \cite{dibra2017human} utilized 3D shape descriptors via shape embedding, Heat Kernel Signatures, and cross-modal Neural Networks for human shape estimation. 
    Sengupta et al. estimated the human shape from the multiple RGB images from different viewpoints by using a probabilistic approach for silhouette representation and joints 2D maps. However, these methods have troubles when applied in reality because of ambiguities in the 2D data \cite{thaler2019influence} like the camera position, resolution, and angle. Moreover, they are limited to only certain resolution images and are hard to be applied on high-resolution images because of a large set of training parameters. Additionally, for anthropometric measurements, Gonzalez et al. \cite{gonzalez2019calvis} proposed a physics simulation method for predicting these measurements. 
    The authors fit the body shape to the template mesh and find inter-joint distances for body part segmentation. 
    Another work \cite{yan2021silhouette} proposed a CNN architecture to extract the heat maps and a regression module to predict the measurements from the heatmaps. However, these anthropometric measurement estimation methods require high computational power for estimation or still work on laboratory-captured images with a single fixed device.

\begin{figure}[t!]
    \centering
    \begin{subfigure}{0.33\linewidth}
		\includegraphics[width=0.95\linewidth]{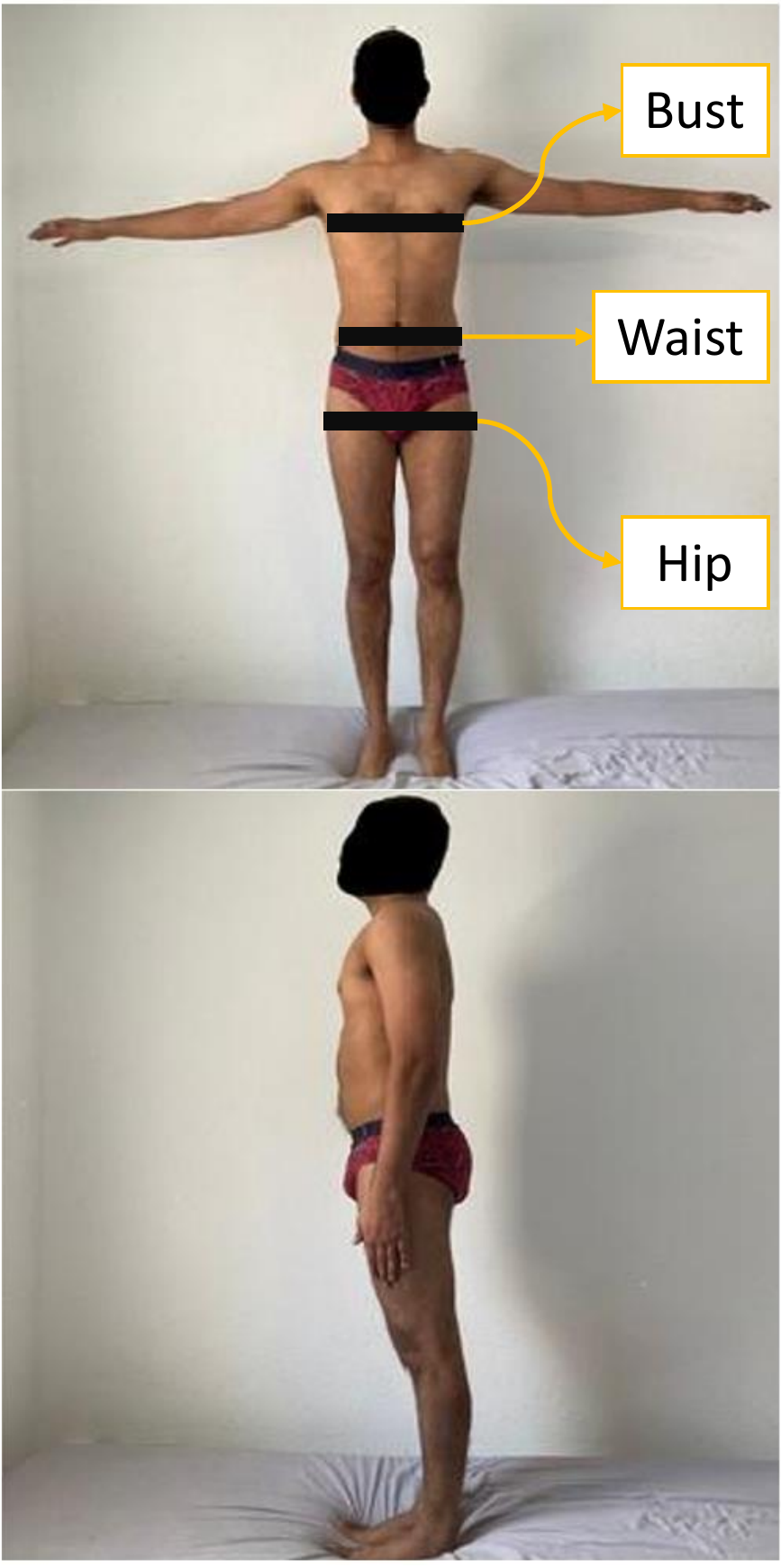}
		\caption{}
		\label{fig:intro_a}
	\end{subfigure}%
	\begin{subfigure}{0.33\linewidth}
		\includegraphics[width=0.97\linewidth]{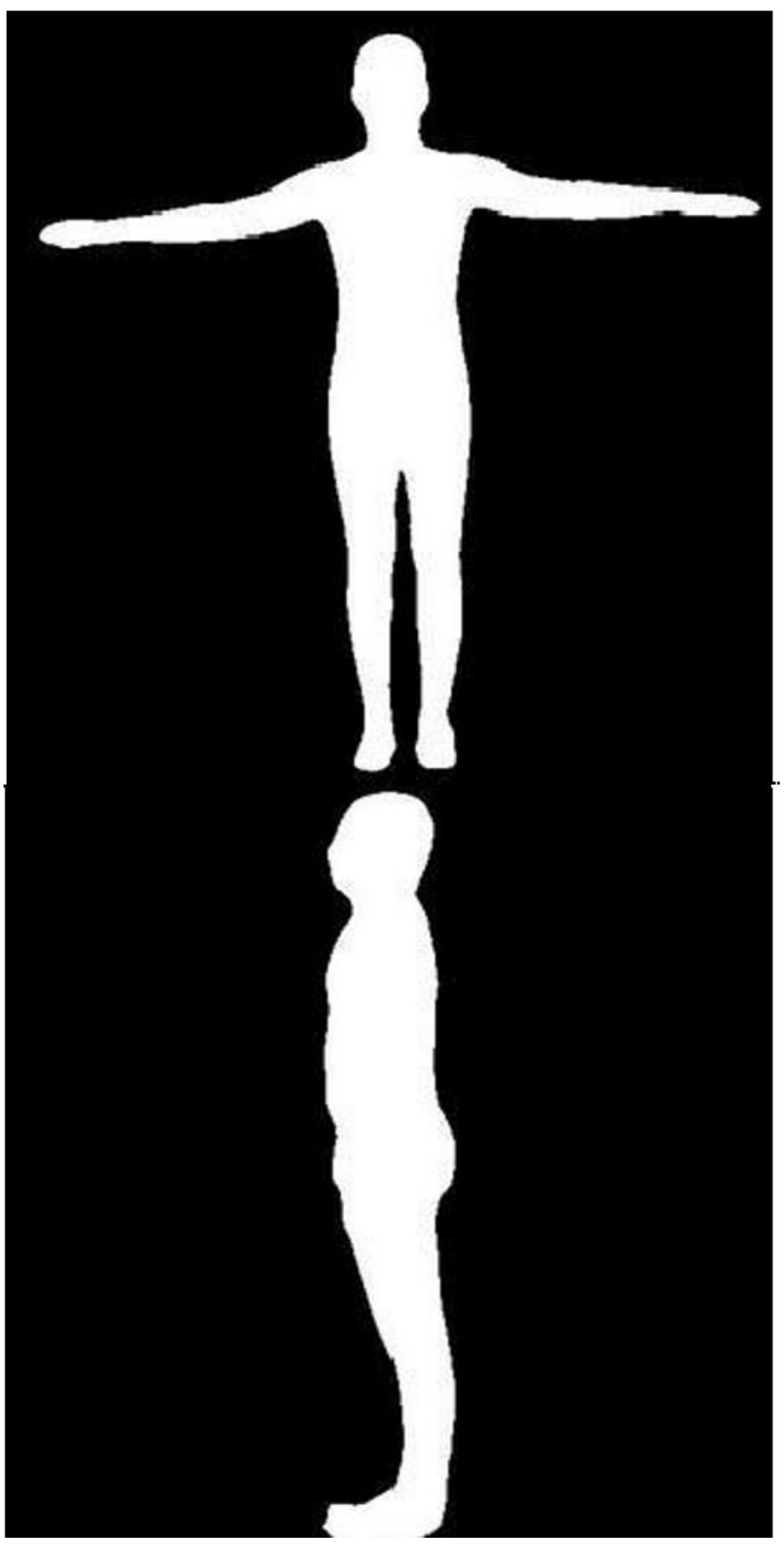}
		\caption{}
		\label{fig:intro_b}
	\end{subfigure}%
	\begin{subfigure}{0.34\linewidth}
		\includegraphics[width=\linewidth]{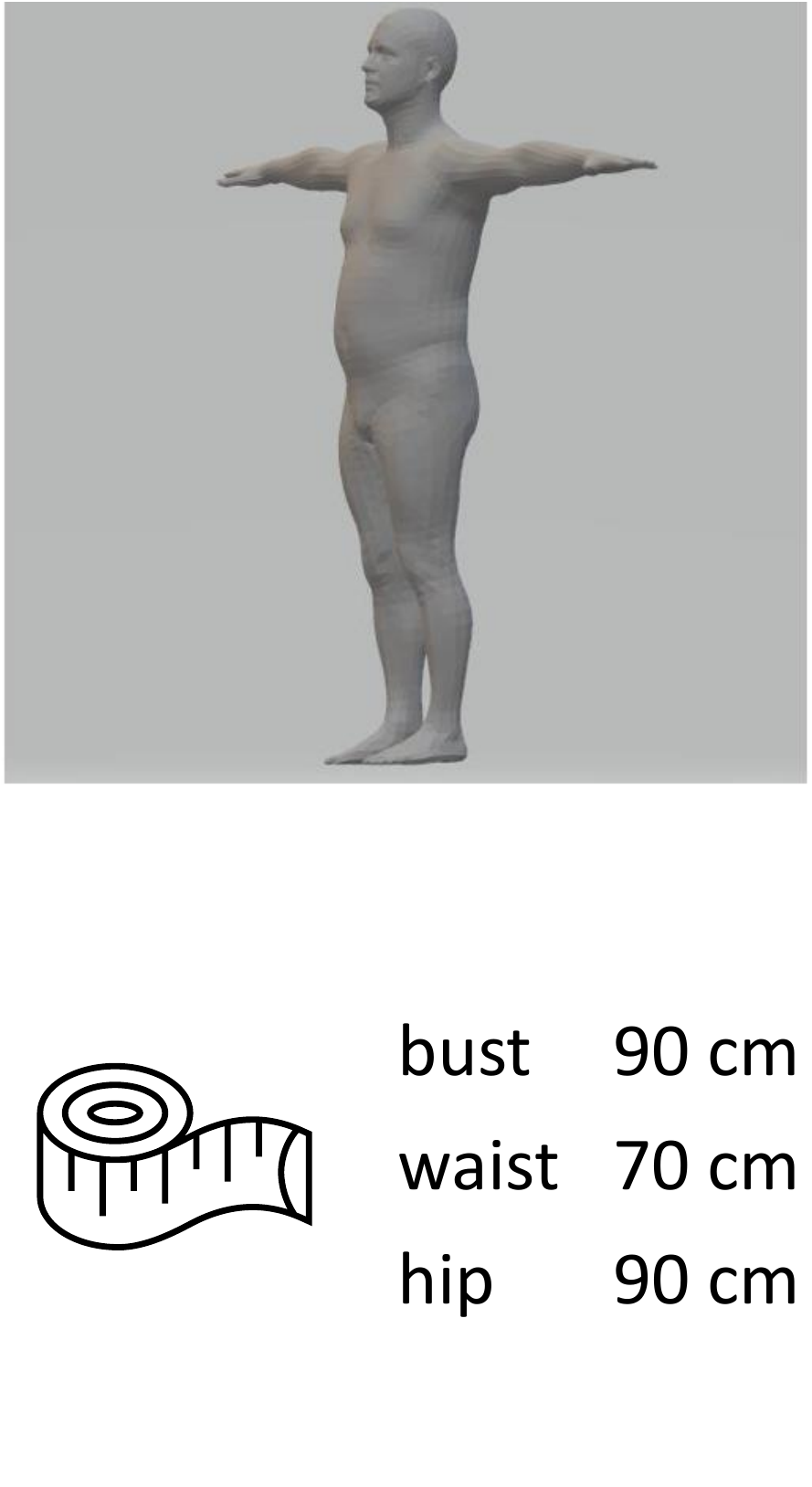}
		\caption{}
		\label{fig:intro_c}
	\end{subfigure}%
     \caption{Input and prediction examples: (a) the frontal and side-view input images, (b) extracted silhouettes, (c) reconstructed 3D body shape and estimated clothing measurements} 
     \label{fig:intro}
    \end{figure}

    To solve the aforementioned problems and overcome the limitations, in this paper, we propose an end-to-end deep neural network architecture for 3D body shape and clothing measurement estimations from the frontal- and side-view images. 
    The examples of the frontal- and side-view images, extracted silhouette images, and the results of the proposed method are shown in \cref{fig:intro}. 
    
    The main contributions of our study can be summarized as follows: (1) To extract human body silhouettes from the frontal- and side-view RGB images, we adopt a U-Net-based \cite{RFB15a} human segmentation model. (2) An auto-encoder is trained to reduce the high-resolution detailed silhouette images of different body shapes to a low-dimensional representation space. (3) A regression module with kernel-based activation maps is proposed to learn the shape parameters for 3D body shape prediction and the bust/waist/hip measurements for clothing fit. (4) The proposed framework is validated with three synthetic and real-world datasets. 

    The rest of the paper is organized as follows. Section \ref{sec:method} provides the details of the proposed method. Section \ref{sec:experimentalresults} presents quantitative experimental results on the three datasets. Finally, Section \ref{sec:conclusion} concludes the paper.

    \begin{figure*}[t!]
    \centering
        \includegraphics[width=0.95\textwidth]{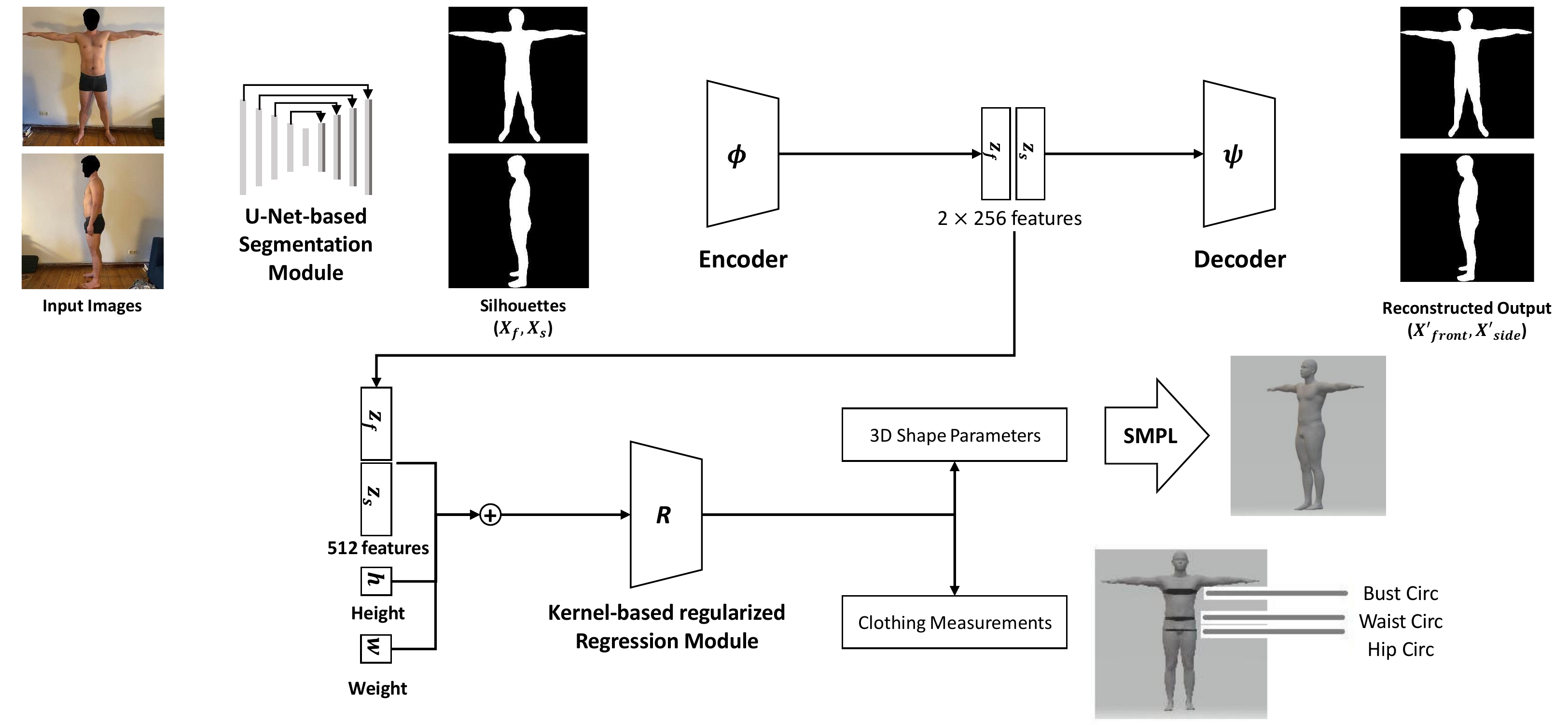}   
     \caption{Overview of the proposed framework:
     The segmentation module extract silhouettes from the two multi-view images. The auto-encoder extracts the embedding features from the silhouette images and the kernel-based regularized regression module predicts the 3D human and the clothing measurements.}
     \label{Fig:architecture}
    \end{figure*}

\section{Proposed Method}
\label{sec:method}

    The proposed framework consists of three phases: A U-Net-based silhouette segmentation module, an auto-encoder to learn the low dimensional feature space of humans, and a kernel-based regularized regression module to estimate the clothing measurements and the 3D body shape through a skinned multi-person linear (SMPL) model \cite{loper2015smpl}. The overview of the proposed framework is shown in \cref{Fig:architecture}, where the input is the two multi-view images of the size of 512 $\times$ 512. The segmentation module extracts the human body silhouette and the auto-encoder learns the 256-dimensional representation for each image. The embedding features with height and weight are then fed into the regression module to estimate the shape and the measurements. We consider height and weight as additional features because learned features are not sufficient to capture the pattern. This is because the pictures can be captured on different mobile phones with different resolution cameras from different distances.

    \subsection{SMPL Model and Training Data}

    In our work, we estimate the 3D body shape by using 
    SMPL \cite{loper2015smpl}, where it factorizes the human body surface into shape ($\beta$) and pose ($\theta$) parameters. The pose($\theta$) is constant, and we deal only with $\beta$ as our problem is more on shape estimation. Thus, the 3D body shape would be: 
    \begin{equation} \label{Eqn1}
        T(\beta) = \widetilde{T} + D_s(\beta)
    \end{equation} 
    where $\widetilde{T}$ denotes the template with zero pose and shape, $D_s(\beta)$ denotes shape dependent deformation, and $T(\beta)$ denotes the deformed vertices.

    We train the proposed architecture on silhouettes of 3D meshes for training in two different views because of the shortage of 2D subjects. We captured the silhouettes, one on the frontal view and the other on the side view. This is done as a scene where the camera is at a fixed distance. Let X $\in$  ${\rm I\!R}^{ 6890 \times 3}$ be the vertices of the 3D human and P be the parameters of the camera. The vertices are transformed on the new camera parameters with an angle of rotation $\theta = 90^{\circ}$.
    \begin{equation}
        \label{eq:rotation}
        X_{new} = P_{\theta = 90^{\circ}} \times X_{old}
    \end{equation}

    We adopted the methods \cite{tejeda2020calvis, STRAPS2020BMVC} for calculating the weight, and bust/waist/hip circumferences measures, which use physics-based simulations by fitting the actual mesh to the SMPL method for computing the measures. We calculated height using cuts between a point A and a point B, and multiply by the separation between cuts. For the weight, we multiplied the volume with the average human body density. The height and weight can be calculated as:
    \begin{equation}
        \label{eq:heightandweight}
        \begin{split}
            h &= N_{cuts} \times D_{cuts}\\
            w &= v \times d \times 1000 ~(\SI{}{\meter\cubed}/\SI{}{\liter})
        \end{split}
    \end{equation}
    where $N_{cuts}$ denotes the number of cuts between the two points, $D_{cuts}$ denotes the separation between two cuts, $v$ and $d$ denote the volume of human mesh and the human body density, respectively, and $d=0.985$ in the experiments.

    \subsection{Auto-encoder}
    In the proposed architecture, the encoder consists of five 3$\times$3 convolutional layers with 32 filters each, followed by batch normalization and Leaky ReLU activation. After every convolution, the output is max-pooled by a 2$\times$2 filter with stride 2. The decoder contains five upsampling layers with a scale factor of 2 and each upsampling layer contains a 3$\times$3 convolutional layer with a batch norm and ReLU activation. In the end, the decoder has a 1$\times$1 convolutional layer with the sigmoid activation squeezing output between 0 and 1. The equations of the encoder and decoder are as follows:
    \begin{equation}
        \label{eq:autoencoder}
        \begin{split}
            z  &=  \phi(X)\\
            X^ \prime  &=  \psi(z)
        \end{split}
    \end{equation}
    where \textit{X} denotes the input silhouette image, $X^ \prime$ is the reconstructed output, $\phi : X \rightarrow z$ is the encoder function that maps input to the low dimensional latent space $z$, and $\psi : z \rightarrow X^ \prime$ denotes the decoder function that reconstructs from $z$. 

    The overall loss function of the auto-encoder for the image reconstruction is defined as follows.
    \begin{equation}
        \label{eq:autoencoder_loss}
        \begin{split}
            \mathop{\mathbb{L}_{rec}} &= \mathop{\mathbb{L}_{BCE}}(X_{f}, X^\prime_{f}) + \mathop{\mathbb{L}_{BCE}}(X_{s}, X^\prime_{s}),\\
            \text{where }&\mathop{\mathbb{L}_{BCE}}= \mathop{\mathbb{E}_{p,q}} [q\log p + (1-q)\log (1-p)],
        \end{split}
    \end{equation}
    where $X_{f}$ and $X_{s}$ denote the silhouettes of the front view and the side view, respectively, and $\mathop{\mathbb{L}_{BCE}}$ denotes the standard binary cross-entropy loss (BCE).

    \subsection{Kernel-based regularized regression module}
     The kernel-based regularized regression module is trained to estimate the 10 Shape parameters and the 3 measurements separately. The estimated 10 shape parameters are then used to reconstruct the 3D skinned human shape in T-Pose using the SMPL method as shown in equation \cref{Eqn1}. Let $z \in \{z_1,  z_2, ... ,  z_{512}\}$ be the extracted images feature vector, h be the height, w be the weight, $\Tilde{Z} \in \{z_1,  z_2, ... ,  z_{512}, h, w\}$ be the input feature vector to the regression module, y be the output shape parameters or the clothing measures, $\Phi_{i}:  \tilde{z}_{i}\rightarrow \Phi(\tilde{z}_{i})$ is the kernel function and $\lambda$ is the regularization constant. Thus, the objective function is given by:
    \begin{equation}
        \label{eq:regressor}
        \begin{split}
        \mathop{\mathbb{L}_{reg}} &= \sum_{i}\Vert y_{i}-W^{T}\Phi_{i}(\tilde{z}_i)\Vert^{2}+\lambda\Vert W\Vert^{2} \\
            \text{where}~  W  &= (\Phi\Phi^{T}+\lambda I)^{-1}\Phi Y^{T}
        \end{split}
    \end{equation}
   
\section{Experimental Results}
    \label{sec:experimentalresults}
    In this section, we discuss the datasets and implementation details and provide the qualitative and quantitative results on different datasets to compare the proposed method with the state-of-the-art methods. 
    
    \textbf{Dataset:} For training and evaluating our model, we have followed the data synthesis method \cite{dibra2016hs} and have labeled the ground truth based on the previous works \cite{tejeda2020calvis, STRAPS2020BMVC}. Since the state-of-the-art methods evaluated on the synthetic data, we show the experimental results on the synthetic data and a 3D scans dataset, named NOMO-3d-400-scans Dataset \cite{yan2020anthropometric}. In addition, we tested our method on real image data from 10 volunteers with RGB images and tape labeled measurements. 

    \textbf{Training:} First, we trained the U-Net-based silhouette segmentation module with four different datasets: Mapillary Vistas \cite{neuhold2017mapillary}, COCO \cite{lin2014microsoft}, Pascal VOC \cite{everingham2010pascal}, and AISegment \cite{AISegment2019}. Second, we trained the auto-encoder module with a batch size of 32 and 50 epochs, using the Adam optimizer with a $0.0001$ learning rate. The performance metric for image reconstruction is average accuracy. 
    Lastly, we trained the kernel-based regularized regression module with the polynomial kernel function, $L_2$ regularization, degree of 3, alpha 0.1, and optimized on $L_2$ loss. The performance metric for clothing measurements is mean absolute error (MAE). The proposed framework was implemented using PyTorch in NVIDIA Tesla V100 DGXS 16GB.
  
    \begin{table}[t!]
        \caption{PCA and AE reconstruction accuracy comparison}
        \label{table:recon}
        \centering
        \begin{tabular}{|l|c|c|c|c|}
        \hline
            \multirow{2}{*}{View}    & \multicolumn{2}{|c|}{PCA} & \multicolumn{2}{|c|}{Auto-encoder} \\ \cline{2-5}
                & Male  & Female    & Male  & Female    \\ \hline
            Front   & 91.00 & 91.20     & 94.98 & 96.21     \\ \hline
            Side    & 99.60 & 99.60     & 95.57 & 96.72     \\ \hline                                 
        \end{tabular}
    \end{table}
 
\begin{figure}[t!]
    \centering
      \includegraphics[width=0.75\linewidth]{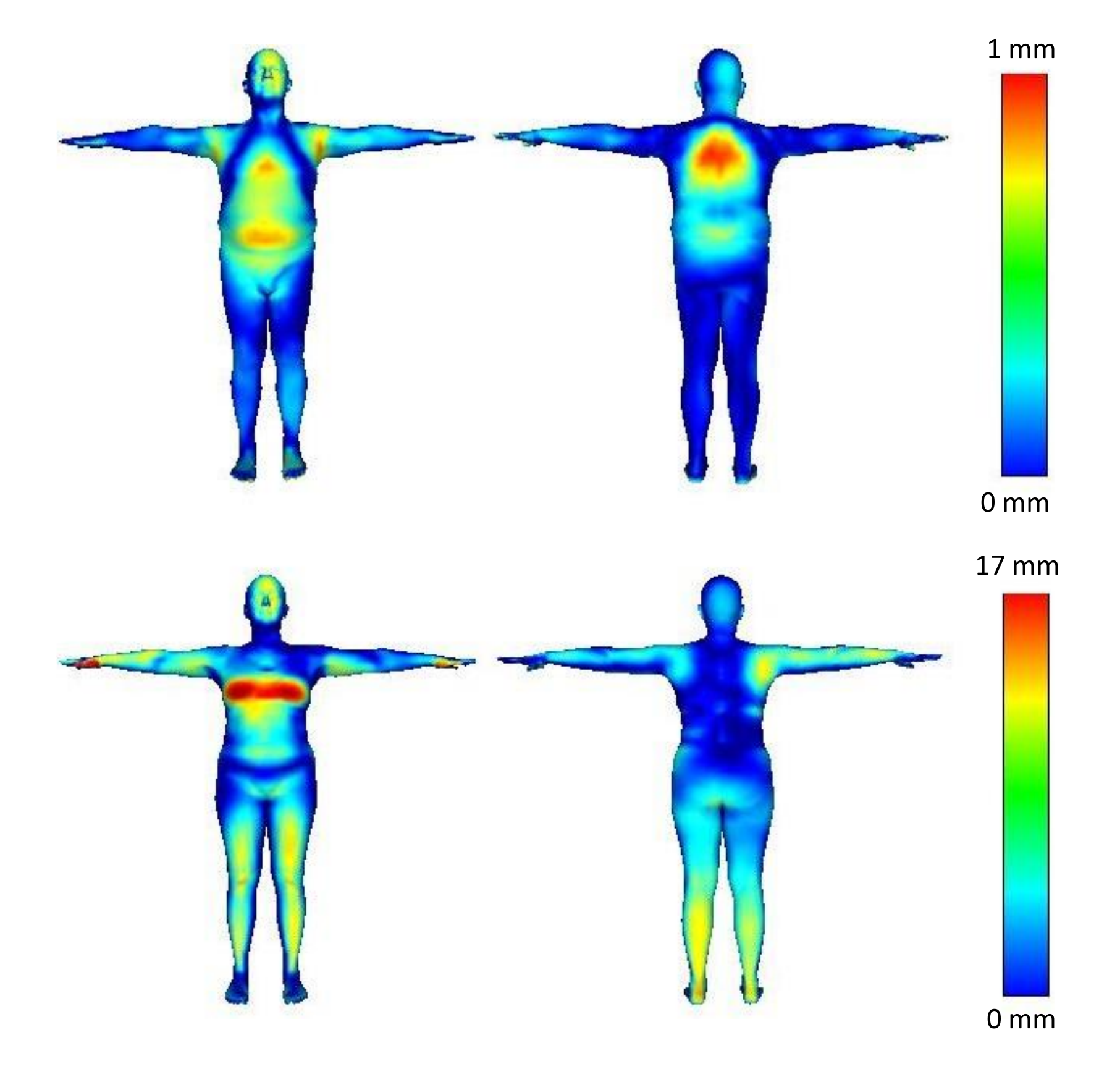}
      \caption{Heatmaps showing the differences between the ground truth and the predicted 3D body shape}
      \label{fig:key}
    \end{figure}

    \textbf{Results:} We show the performance of the proposed method with three different aspects: the performance of reconstruction by the auto-encoder module, the errors of clothing measurement on three different datasets, and the performance of the 3D body shape estimation. First, \cref{table:recon} shows that the comparison reconstruction results with the principal component analysis (PCA) method. The PCA model performed well only on the side-view images but failed on the frontal-view images, but the proposed auto-encoder achieved better accuracy on both views. Second, the comparison results on the synthetic dataset are shown in \cref{table:comparison_synthetic}. The comparison results show that the proposed method provided better measurement performance than other state-of-the-art methods, such as HS-Net \cite{dibra2016hs}, HKS-Net \cite{dibra2017human}, and BoDiEs \cite{vskorvankova2021automatic}. In addition, we evaluated our method on NOMO-3d-400-scans Dataset and the real image data from 10 volunteers. The comparison results are shown in \cref{table:comparison_Nomo} and \cref{table:realdata}. The proposed method gave a better performance on the NOMO-3d-400-scans Dataset than the BODY-rgb method \cite{yan2021silhouette} in terms of three-body measurements. The average error of three-body measurements on the NOMO-3d-400-scans Dataset is \SI{2.83}{\cm} and on the real image data is \SI{2.92}{\cm}.
    Lastly, we used the same training criteria as in the measurements for estimating the shape parameters and reconstructed the 3D body using the SMPL \cite{loper2015smpl} method. We calculated the Hausdorff distance as a per-vertex mean error to compare the distance between the reconstructed mesh and the ground truth. The overall per-vertex mean error is \SI{0.52}{\mm}. The example results are shown in the \cref{fig:key} as heatmaps between the ground truth and the predicted 3D body shape. Our model can estimate 3D shape and clothing measures in 1.4 seconds with a memory size of the proposed model of 25 MB.
    
    \begin{table}[t!]
        \caption{Quantitative body measurement errors on the synthetic dataset (unit: \SI{}{\mm})}
        \label{table:comparison_synthetic}
        \centering
        \resizebox{\columnwidth}{!}{
        \begin{tabular}{|c|c|c|c|c|}
            \hline
            Measures    & HS-Net \cite{dibra2016hs} & HKS-Net \cite{dibra2017human} & BoDiEs \cite{vskorvankova2021automatic} & \textbf{Ours} \\ \hline
            Bust       &  19.10 & 5.60   & 5.24 &   \textbf{5.49}   \\ \hline
            Hip         &  14.90 & 6.90   & 4.92 &   \textbf{2.25} \\ \hline
            Waist       &  18.40 & 7.10   & 3.11 &   \textbf{2.63}   \\ \hline
        \end{tabular}}
    \end{table}

    \begin{table}[!t]
        \caption{Quantitative body measurement errors on the NOMO-3d-400-scans Dataset \cite{yan2020anthropometric} (unit: \SI{}{\mm})}
        \label{table:comparison_Nomo}
        \centering
        \resizebox{\columnwidth}{!}{
        \begin{tabular}{|c|c|c|c|c|}
            \hline
                & \multicolumn{2}{|c|}{Male}    & \multicolumn{2}{|c|}{Female} \\ \hline
            Measures    & BODY-rgb \cite{yan2021silhouette}    & Ours  & BODY-rgb \cite{yan2021silhouette}    & Ours  \\ \hline
            Bust       & 36.10 & 29.95 & 31.70 & 28.79 \\ \hline
            Hip         & 35.50 & 24.81 & 35.50 & 24.00 \\ \hline
            Waist       & 35.30 & 27.88 & 42.70 & 34.45 \\
            \hline
        \end{tabular}}
    \end{table}
    
    \begin{table}[t!]
        \caption{Estimation of three body measurement on 10 human images captured by different cameras (unit: \SI{}{\mm})}
        \label{table:realdata}
        \centering
        \begin{tabular}{|c|c|c|}
            \hline
                        & Male  & Female \\ \hline
                Bust   & 36    & 20    \\  \hline
                Hip     & 18    & 45    \\  \hline
                Waist   & 34    & 22    \\ 
            \hline
        \end{tabular}
    \end{table}
    

\section{Conclusion}
\label{sec:conclusion}
    In this paper, We have presented an efficient and simple architecture that can be adapted by retailers easily. In order to extract silhouettes from the frontal- and side-view images, we adopted the U-Net-based silhouette segmentation module. After extracting the silhouettes, the auto-encoder module extracts the low-dimensional features from the images. Then, the proposed kernel-based regularized regression module predicted accurate 3D body shape by using the SMPL model and three-body measurements without strict camera restrictions. The experimental results showed that the proposed method estimated the body measurements better than the state-of-the-art methods on three different datasets in terms of three different body measurements. In future work, we will collect more real data with more features like body type, age, etc, to improve the performance of the proposed method because the proposed framework was trained on the synthetic dataset only. We believe extending our work by including more features can provide accurate results in measurements estimation.

\section{Acknowledgments}
\label{sec:ack}
The research reported in this paper was supported by the BMBF (German Federal Ministry of Education and Research) in the project VidGenSense (01IW21003). 


\bibliographystyle{IEEEbib}
\bibliography{refs}

\begin{thebibliography}{10}

\bibitem{news}
Biki john,
\newblock ``The size issue – key problem with online clothes shopping,''
\newblock {\em fit analytics}, 2021.

\bibitem{barclay}
Barclay,
\newblock ``Return to sender: Retailers face a ‘phantom economy’ of £7bn
  each year as shopper returns continue to rise,''
\newblock 2018.

\bibitem{dibra2017human}
Endri Dibra, Himanshu Jain, Cengiz Oztireli, Remo Ziegler, and Markus Gross,
\newblock ``Human shape from silhouettes using generative hks descriptors and
  cross-modal neural networks,''
\newblock in {\em Proceedings of the IEEE conference on computer vision and
  pattern recognition}, 2017, pp. 4826--4836.

\bibitem{ma2018learning}
Chao Ma, Yulan Guo, Jungang Yang, and Wei An,
\newblock ``Learning multi-view representation with lstm for 3-d shape
  recognition and retrieval,''
\newblock {\em IEEE Transactions on Multimedia}, vol. 21, no. 5, pp.
  1169--1182, 2018.

\bibitem{ji2018shape}
Zhongping Ji, Xiao Qi, Yigang Wang, Gang Xu, Peng Du, and Qing Wu,
\newblock ``Shape-from-mask: A deep learning based human body shape
  reconstruction from binary mask images,''
\newblock {\em arXiv preprint arXiv:1806.08485}, 2018.

\bibitem{gai20193d}
Zixuan Gai, Xu~Zhao, and Xin Cao,
\newblock ``3d body pose and shape estimation from multi-view images with limb
  geometric constraint,''
\newblock in {\em 2019 IEEE International Conference on Image Processing
  (ICIP)}. IEEE, 2019, pp. 574--578.

\bibitem{yan2021silhouette}
Song Yan, Johan Wirta, and Joni-Kristian K{\"a}m{\"a}r{\"a}inen,
\newblock ``Silhouette body measurement benchmarks,''
\newblock in {\em 2020 25th International Conference on Pattern Recognition
  (ICPR)}. IEEE, 2021, pp. 7804--7809.

\bibitem{thaler2019influence}
Anne Thaler, Sergi Pujades, Jeanine~K Stefanucci, Sarah~H Creem-Regehr, Joachim
  Tesch, Michael~J Black, and Betty~J Mohler,
\newblock ``The influence of visual perspective on body size estimation in
  immersive virtual reality,''
\newblock in {\em ACM Symposium on Applied Perception 2019}, 2019, pp. 1--12.

\bibitem{gonzalez2019calvis}
Yansel Gonzalez-Tejeda and Helmut Mayer,
\newblock ``Calvis: chest, waist and pelvis circumference from 3d human body
  meshes as ground truth for deep learning,''
\newblock in {\em Proceedings of the VIII International Workshop on
  Representation, analysis and recognition of shape and motion FroM Imaging
  data (RFMI 2019)}. ACM, 2019.

\bibitem{RFB15a}
O.~Ronneberger, P.Fischer, and T.~Brox,
\newblock ``U-net: Convolutional networks for biomedical image segmentation,''
\newblock in {\em Medical Image Computing and Computer-Assisted Intervention
  (MICCAI)}. 2015, vol. 9351 of {\em LNCS}, pp. 234--241, Springer,
\newblock (available on arXiv:1505.04597 [cs.CV]).

\bibitem{loper2015smpl}
Matthew Loper, Naureen Mahmood, Javier Romero, Gerard Pons-Moll, and Michael~J
  Black,
\newblock ``Smpl: A skinned multi-person linear model,''
\newblock {\em ACM transactions on graphics (TOG)}, vol. 34, no. 6, pp. 1--16,
  2015.

\bibitem{tejeda2020calvis}
Yansel~Gonzalez Tejeda and Helmut Mayer,
\newblock ``Calvis: chest, waist and pelvis circumference from 3d human body
  meshes as ground truth for deep learning,''
\newblock {\em arXiv preprint arXiv:2003.00834}, 2020.

\bibitem{STRAPS2020BMVC}
Akash Sengupta, Ignas Budvytis, and Roberto Cipolla,
\newblock ``Synthetic training for accurate 3d human pose and shape estimation
  in the wild,''
\newblock in {\em British Machine Vision Conference (BMVC)}, September 2020.

\bibitem{dibra2016hs}
Endri Dibra, Himanshu Jain, Cengiz {\"O}ztireli, Remo Ziegler, and Markus
  Gross,
\newblock ``Hs-nets: Estimating human body shape from silhouettes with
  convolutional neural networks,''
\newblock in {\em 2016 fourth international conference on 3D vision (3DV)}.
  IEEE, 2016, pp. 108--117.

\bibitem{yan2020anthropometric}
Song Yan, Johan Wirta, and Joni-Kristian K{\"a}m{\"a}r{\"a}inen,
\newblock ``Anthropometric clothing measurements from 3d body scans,''
\newblock {\em Machine Vision and Applications}, vol. 31, no. 1, pp. 1--11,
  2020.

\bibitem{neuhold2017mapillary}
Gerhard Neuhold, Tobias Ollmann, Samuel Rota~Bulo, and Peter Kontschieder,
\newblock ``The mapillary vistas dataset for semantic understanding of street
  scenes,''
\newblock in {\em Proceedings of the IEEE international conference on computer
  vision}, 2017, pp. 4990--4999.

\bibitem{lin2014microsoft}
Tsung-Yi Lin, Michael Maire, Serge Belongie, James Hays, Pietro Perona, Deva
  Ramanan, Piotr Doll{\'a}r, and C~Lawrence Zitnick,
\newblock ``Microsoft coco: Common objects in context,''
\newblock in {\em European conference on computer vision}. Springer, 2014, pp.
  740--755.

\bibitem{everingham2010pascal}
Mark Everingham, Luc Van~Gool, Christopher~KI Williams, John Winn, and Andrew
  Zisserman,
\newblock ``The pascal visual object classes (voc) challenge,''
\newblock {\em International journal of computer vision}, vol. 88, no. 2, pp.
  303--338, 2010.

\bibitem{AISegment2019}
AISegment,
\newblock ``Aisegment,''
  \url{https://github.com/aisegmentcn/matting_human_datasets}, 2019.

\bibitem{vskorvankova2021automatic}
Dana {\v{S}}korv{\'a}nkov{\'a}, Adam Rie{\v{c}}ick{\`y}, and Martin Madaras,
\newblock ``Automatic estimation of anthropometric human body measurements,''
\newblock {\em arXiv preprint arXiv:2112.11992}, 2021.

\end{thebibliography}

\end{document}